\documentclass[preprint,review,3p,11pt]{elsarticle} 

\usepackage{amssymb}
\usepackage{algorithm}
\usepackage{algpseudocode} 
\usepackage{booktabs}
\usepackage{threeparttable}
\usepackage{multirow}
\usepackage{amsmath}
\usepackage{setspace}
\usepackage[colorlinks=true, linkcolor=blue, citecolor=blue]{hyperref}
\usepackage[
    labelfont=bf,
    justification=raggedright,
    singlelinecheck=false
]{caption}

\journal{}

\begin{document}

\begin{frontmatter}



\title{CWFBind: Geometry-Awareness for Fast and Accurate Protein-Ligand Docking}

\author[1]{Liyan Jia}
\ead{jialy5@mail2.sysu.edu.cn}
\author[1]{Chuan-Xian Ren\corref{cor1}}
\ead{rchuanx@mail.sysu.edu.cn}
\author[2]{Hong Yan}
\ead{h.yan@cityu.edu.hk}

\affiliation[1]{organization={School of Mathematics, Sun Yat-Sen University},
addressline={135 Xingang West Road},
city={Guangzhou},
postcode={510275},
state={Guangdong},
country={China}}

\affiliation[2]{organization={Department of Electrical Engineering, City University of Hong Kong},
addressline={83 Tat Chee Avenue},
city={Hong Kong},
postcode={999077},
state={Kowloon},
country={China}}

\cortext[cor1]{Corresponding author}

\begin{abstract}
Accurately predicting the binding conformation of small-molecule ligands to protein targets is a critical step in rational drug design. Although recent deep learning-based docking surpasses traditional methods in speed and accuracy, many approaches rely on graph representations and language model-inspired encoders while neglecting critical geometric information, resulting in inaccurate pocket localization and unrealistic binding conformations. In this study, we introduce CWFBind, a weighted, fast, and accurate docking method based on local curvature features. Specifically, we integrate local curvature descriptors during the feature extraction phase to enrich the geometric representation of both proteins and ligands, complementing existing chemical, sequence, and structural features. Furthermore, we embed degree-aware weighting mechanisms into the message passing process, enhancing the model’s ability to capture spatial structural distinctions and interaction strengths. To address the class imbalance challenge in pocket prediction, CWFBind employs a ligand-aware dynamic radius strategy alongside an enhanced loss function, facilitating more precise identification of binding regions and key residues. Comprehensive experimental evaluations demonstrate that CWFBind achieves competitive performance across multiple docking benchmarks, offering a balanced trade-off between accuracy and efficiency. 
\end{abstract}





\end{frontmatter}



\section{Introduction}
\label{sec1}
Biomolecular interactions underpin essential biological processes, with protein–ligand interactions playing a central role in mediating protein function and guiding therapeutic development~\cite{wang2025biomolecular,mou2025deep,ye2022molecular}. These interactions involve the specific, non-covalent binding between small-molecule ligands and macromolecular protein receptors, either in vivo or in vitro~\cite{gainza2023novo}. In modern drug discovery, molecular docking~\cite{muhammed2024molecular,li2021effective} has emerged as a pivotal computational technique to investigate such interactions, addressing the limitations of experimental methods and the vastness of chemical space~\cite{askr2023deep,catacutan2024machine}. By predicting the binding conformation of ligands at atomic resolution, docking offers critical insights into molecular recognition and binding affinity. Beyond its foundational scientific value, molecular docking has become an important tool in drug discovery pipelines, particularly in virtual screening, as it helps accelerate candidate identification and reduce experimental costs~\cite{niazi2023computer}. 

In recent years, molecular docking technologies have undergone remarkable advances, evolving from traditional physics- and chemistry-based simulation software to deep learning–driven predictive frameworks capable of automatically identifying useful patterns from large-scale data. This paradigm shift has significantly improved computational efficiency and automation, offering more powerful tools for investigating protein–ligand interactions. Nevertheless, current research still faces several critical challenges:

\textbf{Insufficient exploitation of 3D structural information in protein–ligand modeling.} Most existing approaches represent proteins and ligands as sequences (e.g., FASTA, SMILES) or graphs, followed by feature extraction using sequence models, language models, or graph neural networks~\cite{shen2016seqkit,arus2020smiles,jiang2025review,pei2023fabind,gao2025fabind+,lu2022tankbind}. Accurate modeling of protein–ligand relationships has been shown to facilitate binding site identification and conformation prediction~\cite{mou2025deep,tubiana2022scannet,gainza2020deciphering}. For ligands, TorchDrug~\cite{zhu2022torchdrug} is commonly used to extract chemical and topological features, while proteins can be modeled using geometric vector perceptrons or graph neural networks incorporating SE(3) equivariant/invariant constraints. A variety of equivariant/invariant graph neural networks have been proposed~\cite{pei2023fabind,gao2025fabind+,lu2022tankbind,zhang2023e3bind,stark2022equibind} and have demonstrated promising performance in molecular docking tasks. However, most existing docking frameworks still fail to fully exploit the spatial geometric information embedded in graph structures, leading to biased modeling of protein–ligand interactions and consequently limiting the accuracy and robustness of binding site identification and docking pose prediction.

\textbf{Difficulty in achieving both high accuracy and high efficiency.} Deep learning–based docking methods can be broadly categorized into: generative model–based methods and regression-based methods. Generative model–based methods, which sample multiple ligand conformations in a generative space, and select the optimal configuration via a confidence model~\cite{corso2023diffdock,yim2024diffusion,plainer2023diffdock}. These methods generally achieve higher accuracy but suffer from low efficiency due to the multi-step sampling and selection process. Regression-based methods, which directly predict the distance matrix~\cite{lu2022tankbind} or atomic coordinates~\cite{stark2022equibind,zhang2023e3bind} of protein–ligand interactions using deep learning models. These methods are computationally efficient but typically lag behind generative methods in accuracy. FABind~\cite{pei2023fabind} strikes a better balance between speed and accuracy by unifying binding pocket prediction and docking within a single framework, thus eliminating reliance on external modules (e.g., P2Rank~\cite{krivak2018p2rank} ) used in methods such as TankBind~\cite{lu2022tankbind} and E3Bind~\cite{zhang2023e3bind}, and reducing training and inference time. However, its binding site stability and docking precision still have room for improvement. Building upon this, FABind+~\cite{gao2025fabind+} was proposed to enhance binding pocket prediction and pose modeling by sampling multiple candidate conformations and selecting the optimal structure, thereby achieving a notable improvement in docking accuracy. However, this improvement in accuracy comes at the expense of increased computational overhead during both training and inference.

In summary, there remains an urgent need for a molecular docking framework capable of efficiently leveraging 3D geometric information while achieving an optimal balance between accuracy and computational efficiency.

To this end, we propose CWFBind, a novel end-to-end docking framework based on a weighted, fast, and accurate docking technique with local curvature feature (LCF). Specifically, LCF is introduced at the protein and ligand representation stage to capture multi-dimensional geometric properties of molecular nodes. These features are subsequently integrated with evolutionary sequence embeddings from ESM-2, the chemical and topological features of TorchDrug, to achieve comprehensive and rich molecular characterization. In the equivariant layer, the message passing mechanism dynamically assigns weights to spatially adjacent atoms, effectively suppressing noise from irrelevant connections and significantly enhancing the model’s ability to distinguish between strong and weak intermolecular interactions. To address class imbalance in pocket prediction, a balanced focal loss is employed, which leverages sample-specific weighting and hard case focusing to optimize performance. Unlike methods that predict multiple potential pockets, this approach identifies a single high-confidence pocket per protein, improving both efficiency and interpretability. Moreover, the pocket radius is dynamically adjusted using a multi-layer perceptron (MLP) conditioned on the ligand’s atomic count, enabling adaptive scaling of the binding region based on ligand size. Finally, the molecular docking stage generates the optimal ligand pose through iterative coordinate refinement. Extensive experiments on the PDBbind v2020 dataset demonstrate that CWFBind outperforms 10 mainstream protein–ligand docking methods in terms of both accuracy and computational efficiency, underscoring its effectiveness and generalization ability.

The main contributions of this paper are summarized as follows:

\begin{itemize}

\item Local curvature information is incorporated into graph node features by leveraging Ollivier’s Ricci curvature as a statistical descriptor. This integration enables the model to precisely capture 3D spatial curvature and structural dependencies within molecular graphs, thereby providing a richer informative representation for conformational prediction in molecular docking.

\item A degree-aware weighting mechanism is introduced, dynamically assigning contribution weights to neighboring atoms based on node degree to enable the capture of hierarchical differences in molecular spatial structure.

\item A balanced focal loss is introduced to address class imbalance in pocket classification, together with an adaptive pocket radius prediction strategy that enhances prediction flexibility and accuracy.

\end{itemize}

The remainder of this paper is organized as follows. Section~\ref{sec2} provides a brief overview of protein–ligand encoding techniques and docking methods. Section~\ref{sec3} describes the proposed CWFBind framework in detail. Section~\ref{sec4} presents the experimental setup and discusses the results. Finally, Section~\ref{sec5} concludes the study and outlines directions for future research.

\section{Related Work}\label{sec2}

\subsection{Protein and ligand representation}
Accurate and informative representation of proteins and ligands is critical for molecular docking tasks, as it directly affects the performance of binding site identification and pose prediction.
Protein and ligand representations are generally categorized into three types: sequence-based representations, structure-based representations, and hybrid representations. Sequence-based representation methods describe proteins in the form of amino acid sequences, such as the FASTA format~\cite{shen2016seqkit}, while common SMILES~\cite{arus2020smiles} strings are used to describe molecular structures.
Structure-based representations incorporate 3D atomic coordinates from formats such as PDB, enabling spatial understanding via grid-based or coordinate-aware models. For example, DCGCN~\cite{sun2025dcgcn} uses two-channel graph convolutional networks to capture spatial relationships between atoms. Similarly, GraphDTA~\cite{nguyen2021graphdta} and EquiBind~\cite{stark2022equibind} employ graph neural networks to model molecular structures and protein–ligand interactions in 3D space. However, these methods often struggle to handle long-range interactions and large conformational flexibility.
Hybrid representations aim to integrate multi-scale features from different modalities. For instance, FlexPose~\cite{dong2023equivariant} fuses 3D protein structures with 2D ligand graphs to build complex interaction models. FABind~\cite{pei2023fabind} and FABind+~\cite{gao2025fabind+} represent proteins and ligands as residue-level and atom-level graphs, respectively, leveraging evolutionary information from ESM-2~\cite{lin2022language} and chemical and topological features from TorchDrug~\cite{zhu2022torchdrug}. These models utilize trans-attention mechanisms and interfacial message-passing modules to capture interaction signals. Despite their strong performance, these frameworks largely overlook the geometric properties embedded in graph structures, limiting their ability to fully model spatial topologies.
Hybrid representations are generally recognized to surpass purely sequence- or structure-based approaches in capturing the intricate patterns of molecular interactions~\cite{hua2025mmdg}. However, achieving effective fusion and balance across diverse information types remains a key challenge. In particular, the explicit encoding of geometric and topological characteristics in molecular graphs is still underdeveloped~\cite{jiang2025traditional,li2025knowledge}. To address this limitation, we incorporate local curvature encoding, enabling a more comprehensive characterization of molecular geometry and interaction landscapes.

\subsection{Pocket prediction}
Binding pocket prediction aims to identify the most likely binding sites on the protein surface before docking. Early geometry-based methods, such as Fpocket~\cite{le2009fpocket}, identify potential cavities by analyzing the protein surface morphology using Voronoi tessellation and $\alpha$-spheres. DoGSiteScorer~\cite{volkamer2012dogsitescorer}, building upon geometric pocket detection, further computes physicochemical descriptors such as hydrophobicity, polarity, and shape to predict binding sites and assess their druggability. In recent years, machine learning, particularly deep learning, has significantly advanced the development of pocket prediction. DeepDrug3D~\cite{pu2019deepdrug3d} employs 3D convolutional neural networks to represent and classify protein–ligand binding sites by transforming biomolecular structures into voxel grids enriched with interaction energy attributes. P2Rank~\cite{krivak2018p2rank} is a machine learning-based tool for protein–ligand binding site prediction that extracts local geometric and physicochemical features from the protein surface and applies a random forest model to score and cluster spatial points, enabling rapid identification of potential binding pockets with a good balance between speed and accuracy. DeepPocket~\cite{aggarwal2021deeppocket} combines geometric methods with deep learning, first detecting candidate binding pockets using Fpocket and then applying 3D convolutional neural networks for rescoring and segmentation, thereby achieving more accurate and generalizable binding site detection. Furthermore, methods such as TankBind~\cite{lu2022tankbind}, E3Bind~\cite{zhang2023e3bind}, and FABind~\cite{pei2023fabind} can automatically learn complex patterns of binding sites and ligand conformations from protein structures in an end-to-end manner. However, they typically rely on predefined fixed-size pocket regions, which limits their ability to adapt to diverse protein structures. 

\subsection{Molecular docking}
Molecular docking predicts the optimal binding pose of a ligand to a protein, typically by searching over possible conformations and scoring their binding affinity. A variety of docking tools have been developed to address this challenge, including GLIDE~\cite{friesner2004glide}, VINA~\cite{trott2010autodock}, SMINA~\cite{koes2013lessons}, and GNINA~\cite{mcnutt2021gnina}. These tools are widely adopted in both academic and industrial settings due to their distinct advantages. GLIDE~\cite{friesner2004glide}, part of the Schrödinger suite, is a high-precision docking platform offering multiple modes such as SP and XP, and supports induced fit docking through integration with the Prime module, making it well-suited for precision screening. VINA~\cite{trott2010autodock}, an open-source tool, is extensively used in large-scale virtual screening owing to its speed and practical scoring function. SMINA~\cite{koes2013lessons}, as an extension of VINA, expands support for molecular formats, optimizes docking algorithms for flexible molecules, and offers customizable parameter settings. In contrast, GNINA~\cite{mcnutt2021gnina} introduces deep learning by incorporating convolutional neural networks to enhance scoring accuracy and supports multi-GPU parallelism, marking a significant advancement in data-driven docking approaches. 

Recent advances in deep learning have inspired two major classes of docking approaches: generative model–based and regression-based. 
Generative model–based methods explore ligand conformations within a learned generative space and select optimal poses using a confidence model. DiffDock~\cite{corso2023diffdock} pioneered the formulation of ligand conformation prediction as a generative modelling task by mapping ligand poses onto a non-Euclidean manifold and decoupling translational, rotational, and torsional degrees of freedom via a diffusion process. Through progressive sampling, DiffDock generates flexible ligand binding conformations, enabling the exploration of multimodal binding modes from a global perspective. DiffDock-Pocket~\cite{plainer2023diffdock}, an extension of DiffDock, adapts the framework for docking within predefined binding pockets. It simultaneously optimizes ligand poses and protein side-chain conformations by defining a diffusion process over both ligand configurations and protein side-chain torsion angles.  Conversely, diffusion-based molecular docking methods often suffer from computational inefficiency due to their reliance on multi-step sampling processes and exhibit limited generalizability as a result of biases in the training data.

In contrast, regression-based methods are more efficient as they directly predict spatial relationships in protein-ligand complexes. TankBind~\cite{lu2022tankbind} employs a trigonometry-aware neural network to predict a protein–ligand distance map, which is then converted into a docking conformation via gradient descent, with training driven by a weighted distance loss. E3Bind~\cite{zhang2023e3bind} introduces an end-to-end E(3)-equivariant architecture that incorporates geometric constraints and local binding context, iteratively refining ligand coordinates during docking. FABind~\cite{pei2023fabind} further integrates binding site identification and complex conformation prediction into an enhanced E(3)-equivariant framework, jointly optimising pocket classification and docking through a multi-distance loss. Building on this, FABind+~\cite{gao2025fabind+} improves ligand conformational plausibility through an alignment loss, enhances site exploration with a confidence-guided sampling strategy, and incorporates dynamic pocket radius prediction to better adapt to structural variability. 
Despite these advances, such end-to-end approaches still face challenges in generalising to unseen protein families and handling highly irregular binding environments.

\section{Method}\label{sec3}

\begin{figure}[t]  
    \centering
    \includegraphics[width=\textwidth]{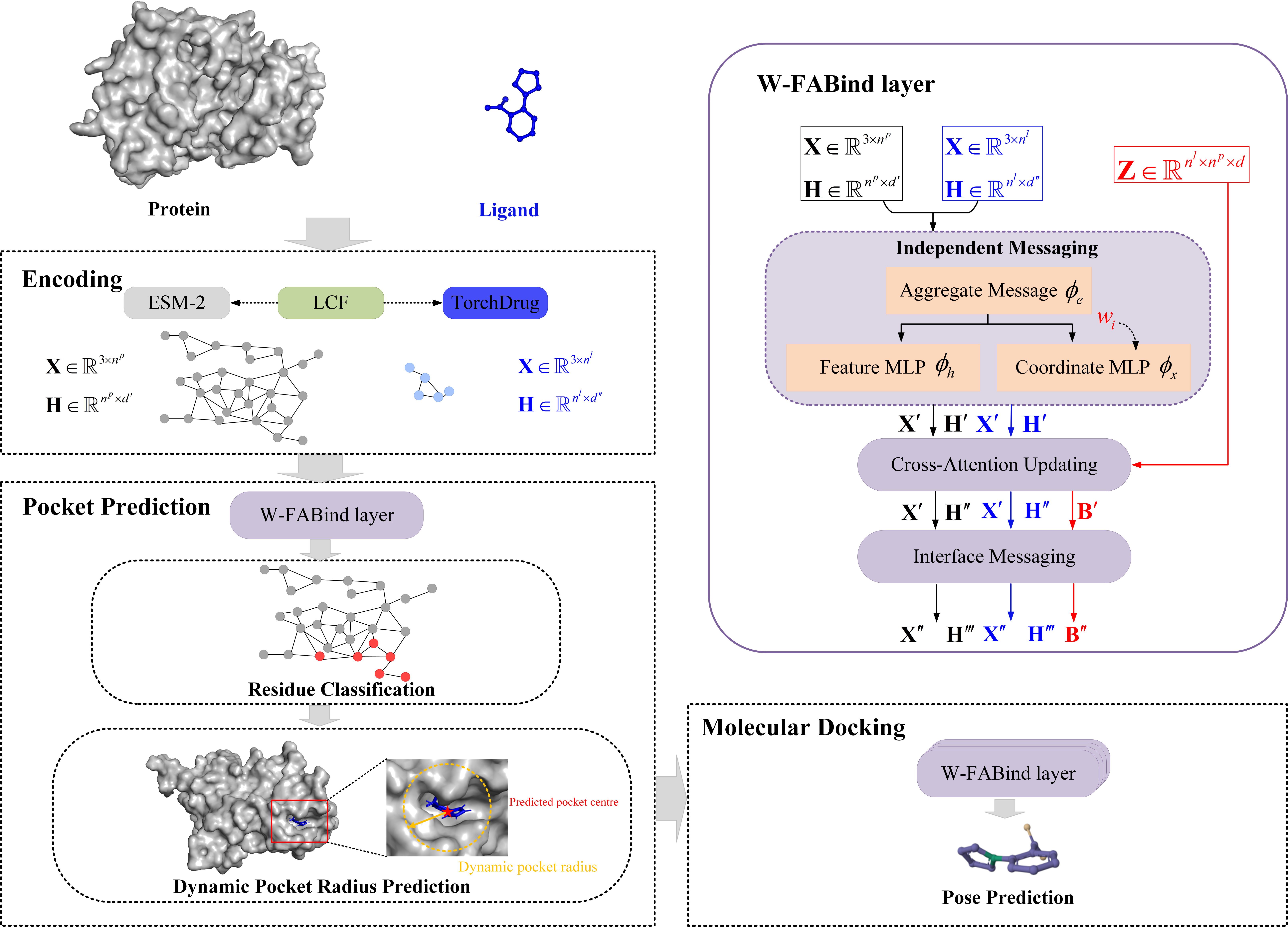} 
    \caption{Framework of CWFBind. Left: First, input the entire protein and ligand and encode them, including ESM-2, TorchDrug, and LCF. The pocket prediction module then updates features via the CWFBind layer, classifies residues to identify pocket sites, and calculates the pocket center and radius. Finally, the docking module iteratively moves the ligand to the pocket centre through four CWFBind layers and returns the predicted ligand pose. Top right: CWFBind layer structure, with each layer containing three modules: independent message passing, cross-attention update, and interface message passing, which update the coordinates and representations of nodes.} 
    \label{figure1}
\end{figure}

The proposed CWFBind framework is illustrated in Fig.~\ref{figure1}. Section~\ref{subsec3.1} introduces the notations used throughout the paper. The encoding strategies for proteins and ligands are detailed in Section~\ref{subsec3.2}. Section~\ref{subsec3.3} presents the core architecture of CWFBind, followed by a description of the pocket prediction module in Section~\ref{subsec3.4} and the docking prediction process in Section~\ref{subsec3.5}. Finally, Section~\ref {subsec3.6} provides the training process and algorithm pseudo-code.

\subsection{Notations}\label{subsec3.1}

We define the ligand–protein complex graph as $\mathcal{G} = \{ \mathcal{V} := (\mathcal{V}^l, \mathcal{V}^p), \mathcal{E} := (\mathcal{E}^l, \mathcal{E}^p, \mathcal{E}^{lp}) \}$, where $\mathcal{V}$ and $\mathcal{E}$ represent the sets of nodes and edges, respectively.  In the ligand subgraph  $\mathcal{G}^l = \{ \mathcal{V}^l, \mathcal{E}^l \}$, each node $v_i = (\mathbf{h}_i, \mathbf{x}_i) \in \mathcal{V}^l$ corresponds to an atom, where $\mathbf{h}_i \in \mathbb{R}^{d_1}$ denotes the atom-level feature vector and $\mathbf{x}_i \in \mathbb{R}^3$ denotes the atom's 3D coordinate. The total number of ligand atoms is denoted as $n^l$. The edge set $\mathcal{E}^l$ encodes the chemical bonds within the ligand. In the protein subgraph $\mathcal{G}^p = (\mathcal{V}^p, \mathcal{E}^p)$, each node $v_j = (\mathbf{h}_j, \mathbf{x}_j) \in \mathcal{V}^p$ represents an amino acid residue, where $\mathbf{h}_j \in \mathbb{R}^{d_2}$ denotes the residue-level feature vector, and $\mathbf{x}_j \in \mathbb{R}^3$ corresponds to the 3D coordinate of the residue’s $C_\alpha$ atom. The number of residues is defined as $n^p$. Edges within $8 \text{\AA}$ of the node distance are selected to construct the edge set $\mathcal{E}^p$. In addition, $\mathcal{E}^{lp}$ denotes the set of edges connecting nodes in ligands and proteins, which is composed of all edges with distances between nodes less than $10 \text{\AA}$ in $\mathcal{V}^l$ and $\mathcal{V}^p$. For clarity, the indices $i, k$ and $j, k'$ are used to refer to ligand and protein nodes, respectively.

\subsection{Encoding of protein and ligand}\label{subsec3.2}

Consistent with previous studies, protein and ligand are encoded as graph structures. Specifically, ligand node feature is precomputed using TorchDrug~\cite{zhu2022torchdrug}, yielding a 52-dimensional vector for each atom. This vector encodes fundamental atomic properties and chemical information. For protein nodes, contextual features are extracted using the pre-trained ESM-2 model~\cite{lin2022language}, a 33-layer Transformer with 650 million parameters, which captures long-range dependencies between amino acid residues via self-attention mechanisms. Given an amino acid sequence as input, ESM-2 produces a 1280-dimensional embedding for each residue, encoding its structurally and chemically relevant contextual information.

However, relying solely on the ESM-2 and TorchDrug toolkits is insufficient for fully characterizing proteins and ligands, as both lack explicit encoding of the geometric structure of molecular graphs. ESM-2 focuses exclusively on amino acid sequences, overlooking critical 3D conformational details, while TorchDrug-derived ligand features fail to capture long-range dependencies and higher-order structural motifs. In protein–ligand systems, the compatibility of binding sites and ligands is not only determined by their shapes but also by the way local structures curve and connect within the molecular graph. To better capture this interplay, we incorporate LCF computed via ORC. ORC quantifies how neighborhood structures around connected nodes differ, effectively measuring the “connective tightness” and structural robustness of molecular graphs. Previous studies have demonstrated that graph neural networks enhanced with discrete Ricci curvature—whether Ollivier’s or Forman’s—achieve superior expressive power by integrating both topological and geometric information~\cite{shen2024curvature,wee2021forman,wee2021ollivier}. This allows curvature-based features to reveal regions of high geometric variability, such as pocket entrances or flexible ligand fragments, that are critical for binding adaptability. Moreover, unlike raw 3D coordinates, which are sensitive to alignment and conformational noise, or simple graph statistics, which overlook fine-grained geometric relationships, ORC provides an intrinsically geometry-aware and topology-integrated measure that remains stable under small perturbations. This enables a richer joint representation of proteins and ligands, complementing sequence- and topology-only features and ultimately enhancing docking performance.

\begin{figure}[t]%
\centering
\includegraphics[width=\textwidth, trim = 0cm 4.3cm 0cm 1.0cm clip]{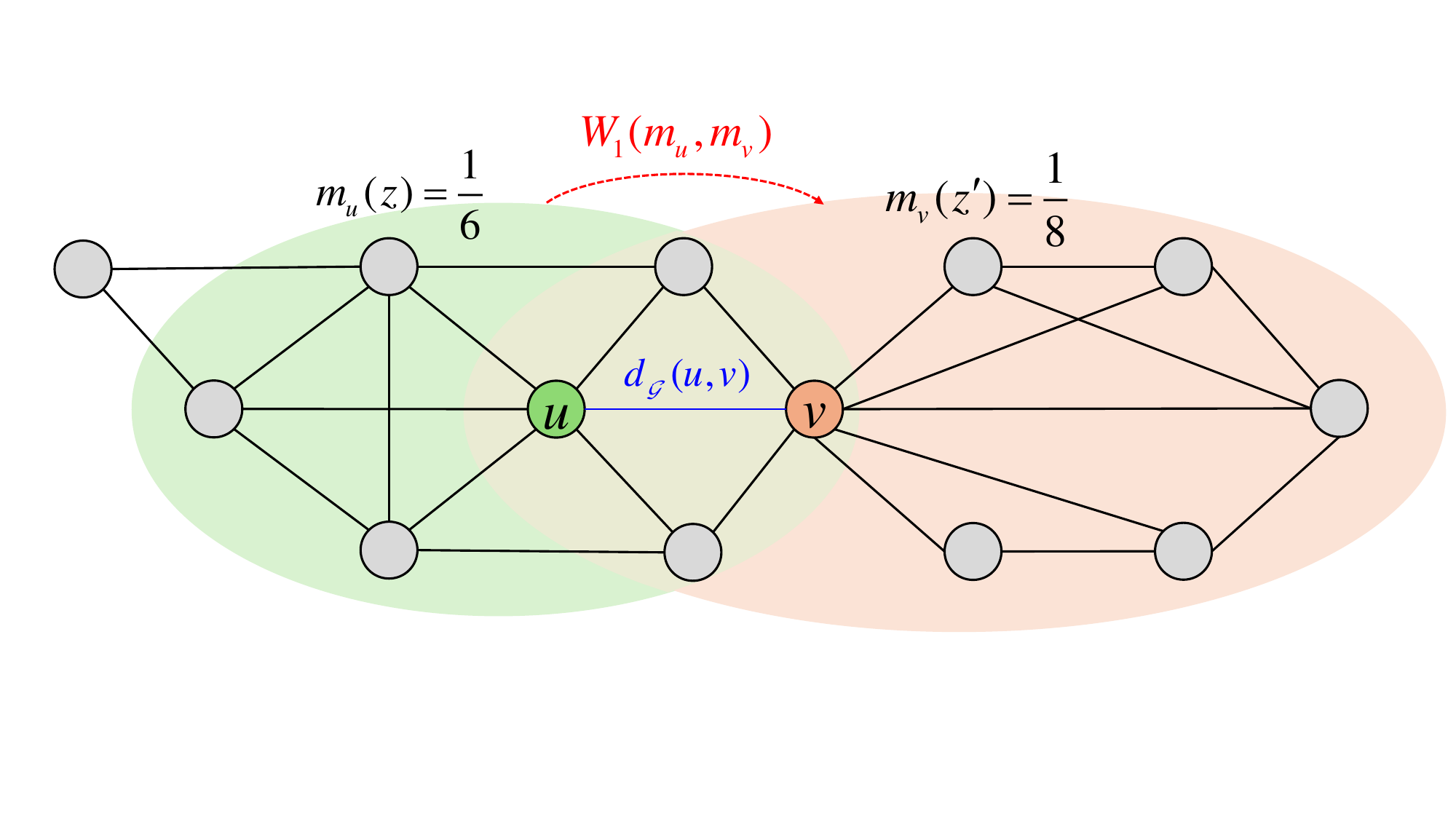}
\caption{The ORC corresponding to the edge \((u, v)\).}
\label{figure2}
\end{figure}

LCF is based on ORC, which is an innovative way to integrate the information of the local curvature distribution of a graph into the node features. Specifically, ORC is a concept that integrates Ricci curvature with optimal transport theory to portray the geometric properties of graph structures. For adjacent vertices \( u \) and \( v \) in the graph \( \mathcal{G} = (\mathcal{V}, \mathcal{E}) \), we define a uniform measure \( m_i \) over the one-hop neighbourhood \( \mathcal{N}(i) \), where for each neighbour \( z \in \mathcal{N}(i) \), the measure is given by $m_i(z) := \frac{1}{\deg(i)}$.
Here, \( i \in \{u, v\} \), and \( \deg(i) \) denotes the degree of vertex \( i \). The transport cost between two vertex neighbours is measured by the Wasserstein-1 distance

\begin{equation}
W_1(m_u, m_v) = \inf_{m \in \Gamma(m_u, m_v)} \int_{(z,z') \in \mathcal{V} \times \mathcal{V}} d(z, z') m(z, z') dz dz',
\end{equation}
where $\Gamma(m_u, m_v)$ denotes the set of all joint measures on $\mathcal{V} \times \mathcal{V}$ with $m_u$ and $m_v$ as edge measures. On this basis, ORC is defined as:

\begin{equation}
\kappa(u, v) := 1 - \frac{W_1(m_u, m_v)}{d_\mathcal{G}(u, v)},
\end{equation}
where $d_\mathcal{G}(u, v)$ is the distance between vertices $u$ and $v$ in the graph $\mathcal{G}$. Since $u$ and $v$ are adjacent, we have $d_\mathcal{G}(u, v) = 1$. The ORC quantifies the geometric properties of the graph by computing the optimal transport cost $W_1(m_u, m_v)$ between the probability measures $m_u$ and $m_v$. A curvature value $\kappa(u, v)$ close to 1 indicates that the neighbourhood structures of $u$ and $v$ are highly similar, suggesting a locally flat graph structure. Conversely, a smaller $\kappa(u, v)$ implies greater dissimilarity between the neighbourhoods and indicates a more curved or heterogeneous local geometry. The ORC simplified graph for edge \((u, v)\) is shown in Fig.~\ref{figure2}.

For a graph $\mathcal{G} = (\mathcal{V}, \mathcal{E})$, where $\mathcal{V}$ and $\mathcal{E}$ denote the node and edge sets, respectively, the curvature multiset associated with a node $v \in \mathcal{V}$ is defined as $\text{CMS}(v) = \{ \kappa(u, v) : (u, v) \in \mathcal{E} \}$, where $\kappa(u, v)$ denotes the ORC of edge $(u, v)$. $\text{LCF}(v)$ consists of five statistics of $\text{CMS}(v)$:

\begin{equation}
\begin{aligned}
\text{LCF}(v) = [ &\min(\text{CMS}(v)), \max(\text{CMS}(v)), \text{mean}(\text{CMS}(v)), \\
                & \text{std}(\text{CMS}(v)), \text{median}(\text{CMS}(v))].
\end{aligned}
\end{equation}

In addition, we adopt an outer product modelling (OPM)~\cite{zhang2023e3bind} scheme to construct pairwise embeddings between protein residues and ligand atoms. For each pair $(i, j)$, the interaction feature is defined as $z_{ij} = \text{Linear}(\text{Linear}(\mathbf{h}_i)\otimes \text{Linear}(\mathbf{h}_j))$ enabling explicit encoding of cross-molecular dependencies through bilinear feature fusion.

\subsection{CWFBind Layer}\label{subsec3.3}

Each FABind layer~\cite{pei2023fabind} consists of three key steps: independent messaging, cross-attention updating, and interface messaging. The independent messaging layer updates node coordinates using uniform weights, assuming equal influence from all neighboring nodes. This assumption fails to reflect real-world variations in chemical context, spatial proximity, and informational relevance among neighbors. To better capture these distinctions, we introduce a degree-based weighting mechanism that leverages the node degree as a proxy for atomic importance. High-degree atoms are typically located at structurally or functionally important sites and carry more informative cues for molecular interactions. By incorporating node degree into the message-passing process, the model can dynamically adjust the influence of neighboring atoms, thereby enhancing its sensitivity to local structural variations. The cross-attention and interface messaging components remain consistent with the original FABind architecture. The process of passing ligand node information is described in more detail below, and the protein is updated in a similar way.

\subsubsection{Independent Messaging}
It is a process of transferring information within the ligand via an equivariant graphic convolutional layer. Let $\mathbf{h}_k^l$ denote the feature of the $k$-th nearest neighbor of the ligand node $\mathbf{h}_i^l$ at the $l$-th layer. The message $\mathbf{m}_{ik}$ from node $k$ to node $i$ is computed using a MLP $\phi_e$. Subsequently, the node feature $\mathbf{h}_i^{l+1}$ and coordinate $\mathbf{x}_i^{l+1}$ are updated through independent message passing.
\begin{equation} \label{eq:update}
\begin{aligned}
\mathbf{m}_{ik} &= \phi_e \left( \mathbf{h}_i^l, \mathbf{h}_k^l, \| \mathbf{x}_i^l - \mathbf{x}_k^l \|^2 \right), \\
\mathbf{h}_i^{l+1} &= \mathbf{h}_i^l + \phi_h \left( \mathbf{h}_i^l, \sum_{k \in \mathcal{N}(i \mid \mathcal{E}^l)} \mathbf{m}_{ik} \right), \\
\mathbf{x}_i^{l+1} &= \mathbf{x}_i^l + \sum_{k \in \mathcal{N}(i \mid \mathcal{E}^l)} \mathbf{w}_{ik} ( \mathbf{x}_i^l - \mathbf{x}_k^l ) \phi_x ( \mathbf{m}_{ik} ), \\
\mathbf{w}_{ik} &= \mathbf{d}_k^l \bigg/ \sum_{k \in \mathcal{N}(i \mid \mathcal{E}^l)} \mathbf{d}_k^l,
\end{aligned}
\end{equation}
where $\phi_h$ and $\phi_x$ are the MLP. $\mathcal{N}(i | \mathcal{E}^l)$ denotes the neighbours of node $i$ with respect to ligand interior edge $\mathcal{E}^l$. $\mathbf{w}_{ik}$ is the weight of the degree of the $k$-th nearest neighbour of the node $i$, and $\mathbf{d}_k^l$ denotes the degree of node $k$. By incorporating degree-based weighting, the model can selectively emphasize more informative neighbors during coordinate updates, leading to a more accurate representation of intramolecular interactions.

\subsubsection{Cross-Attention Updating}

This step involves performing cross-attention updates on all protein/ligand nodes to capture protein-ligand interactions. Specifically, \(\mathbf{q}_i^{(h)}\), \(\mathbf{k}_j^{(h)}\), and \(\mathbf{v}_j^{(h)}\) are linear projections of the node embeddings. \(b_{ij}^{(h)} = \text{Linear}(\mathbf{z}_{ij}^l)\) denotes a linear transformation applied to the protein–ligand pair embedding \(\mathbf{z}_{ij}^l\), where concat refers to the vector concatenation operation.

\begin{equation}
\begin{aligned}
a_{ij}^{(h)} &= \text{softmax}_j \left( \frac{1}{\sqrt{c}} \mathbf{q}_i^{(h) \top} \mathbf{k}_j^{(h)} + b_{ij}^{(h)} \right), \\
\mathbf{h}_i^{l+1} &= \mathbf{h}_i^l + \text{Linear} \left( \text{concat}_{1 \leq h \leq H} \left( \sum_{j = 1}^{n^{p*}} a_{ij}^{(h)} \mathbf{v}_j^{(h)} \right) \right)
\end{aligned}
\end{equation}

Further update the pair embedding, \(\mathbf{z}_{ij}^{l+1} = \text{OPM}(\mathbf{h}_i^{l+1}, \mathbf{h}_j^{l+1})\), based on the updated node embeddings, \(\mathbf{h}_i^{l+1}\) and \(\mathbf{h}_j^{l+1}\).

\subsubsection{Interface Messaging}

Collaborative updating of node features and coordinates at protein-ligand contact interfaces via geometry-aware messaging with attentional bias to accurately portray dynamic conformational rearrangements of interfacial regions during binding. Firstly, for the external edge $(i,j) \in \mathcal{E}^{lp*}$, the mapping of multimodal information is achieved using MLP, i.e., \(\mathbf{q}_i = \phi_q(\mathbf{h}_i^l)\), \(\mathbf{k}_{ij} = \phi_k(\|\mathbf{x}_i^l - \mathbf{x}_j^l\|^ 2, \mathbf{h}_j^l)\), \(\mathbf{v}_{ij} = \phi_v(\|\mathbf{x}_i^l - \mathbf{x}_j^l\|^2, \mathbf{h}_j^l)\), \(b_{ij} = \phi_b(\mathbf{z}_{ ij } ^l)\). The attention weight \(\alpha_{ij}\) is then calculated, which in turn updates the node features \(\mathbf{h}_i^{l + 1}\) and coordinates \(\mathbf{x}_i^{l + 1}\).

\begin{equation}
\begin{aligned}
\alpha_{ij} &= \frac{\exp(\mathbf{q}_i^{\top} \mathbf{k}_{ij} + b_{ij})}{\sum_{j \in \mathcal{N}(i | \mathcal{E}^{lp*})} \exp(\mathbf{q}_i^{\top} \mathbf{k}_{ij} + b_{ij})}, \\
\mathbf{h}_i^{l + 1} &= \mathbf{h}_i^l + \sum_{j \in \mathcal{N}(i | \mathcal{E}^{lp*})} \alpha_{ij} \mathbf{v}_{ij}, \\
\mathbf{x}_i^{l + 1} &= \mathbf{x}_i^l + \sum_{j \in \mathcal{N}(i | \mathcal{E}^{lp*})} \alpha_{ij} (\mathbf{x}_j^l - \mathbf{x}_i^l) \phi_{xv}(\mathbf{v}_{ij}),
\end{aligned}
\end{equation}
where \(\phi_b\) and \(\phi_{xv}\) are MLPs.

\subsection{Pocket Prediction}\label{subsec3.4}

In molecular docking, pocket prediction is formulated as a binary classification task at the residue level to identify protein binding sites. Considering the class imbalance issue between pocket residues and non-pocket residues, we adopt an improved version of the focal loss~\cite{lin2017focal} with a sample-specific weighting scheme. Classification is performed based on the updated protein-ligand map of the $M_1$ layer of CWFBind. The loss is defined as:

\begin{equation}
\begin{aligned}
\mathcal{L}_{cls} = \frac{1}{N} \sum_{i=1}^{N} \frac{n_i^l}{n_i^p} \Bigg\{ &- \sum_{j=1}^{n_i^l} \Big[ y_j (1 - \hat{y}_j)^{\gamma} \log(\hat{y}_j) + (1 - y_j) \hat{y}_j^{\gamma} \log(1 - \hat{y}_j) \Big] \Bigg\},
\end{aligned}
\end{equation}
where $N$ denotes the total number of protein-ligand complexes in the training set. $n_i^l$ is the number of residues in the $i$-th protein, and $n_i^p$ is the number of pocket residues within that protein. The outer weighting factor $\frac{n_i^l}{n_i^p}$ increases the contribution of proteins with fewer pocket residues, compensating for imbalance across complexes. $y_j \in \{0, 1\}$ is the ground-truth label for whether residue $j$ is part of the binding pocket, and $\hat{y}_j \in [0, 1]$ is the predicted probability. The focusing parameter $\gamma$ controls the down-weighting of well-classified samples. A higher $\gamma$ (set to 2 in the experiment) increases the model’s focus on hard-to-classify residues. The inner term $y_j (1 - \hat{y}_j)^{\gamma} \log(\hat{y}_j) + (1 - y_j) \hat{y}_j^{\gamma} \log(1 - \hat{y}_j)$ reduces the loss contribution from well-classified samples and retains a higher penalty for hard or misclassified ones. 

In molecular docking tasks, accurately determining pocket centres is critical to achieving a deeper understanding of protein-ligand interactions. Traditional direct discrete selection suffers from the non-microscopicity problem, which contradicts backpropagation optimisation. We use probability-weighted averaging to calculate the pocket centre.
\begin{equation}
\mathbf{\hat{x}}^p = \frac{1}{n^{p*}} \sum_{j=1}^{n^{p*}} \gamma_j^p \mathbf{x}_j^p,
\end{equation}
where $\mathbf{\hat{x}}^p$ is the final predicted pocket centre coordinate, and $n^{p*}$ is the number of pocket residues. $\gamma_j^p$ is the weight of each predicted pocket residue $j$, computed by Gumbel-Softmax~\cite{jang2017categorical}, which reflects the probability that the residue belongs to the pocket, allowing for a differentiable approximation of the discrete selection process, and $\mathbf{x}_j^p$ are the coordinates of the predicted pocket residue $j$.

To improve the accuracy of binding site prediction, we introduce a spatial constraint that penalizes deviations between the predicted pocket center \(\mathbf{\hat{x}}^p\) and the ground truth center \(\mathbf{x}^{p}\). This constraint leverages the Huber loss~\cite{huber1992robust} to balance sensitivity to small errors and robustness to outliers, particularly important for noisy or flexible binding regions. 
\begin{equation}
\begin{aligned}
\mathcal{L}_{\text{cen}} &= l_{Huber}(\hat{\mathbf{x}}^p, \mathbf{x}^p) \\
&=
\begin{cases}
\displaystyle \frac{1}{2} \left\| \hat{\mathbf{x}}^p - \mathbf{x}^p \right\|_2^2, & \text{if } \left\| \hat{\mathbf{x}}^p - \mathbf{x}^p \right\|_2 \leq \delta, \\[8pt]
\delta \left( \left\| \hat{\mathbf{x}}^p - \mathbf{x}^p \right\|_2 - \frac{1}{2}\delta \right), & \text{otherwise},
\end{cases}
\end{aligned}
\end{equation}
where $\delta > 0$ is a threshold hyperparameter controlling the transition between quadratic and linear penalty regimes.

The preliminary predicted radius \(\hat{r}\) is obtained from the output of an MLP regression head. To better align the pocket size with the spatial requirements of the ligand, we incorporate an adjustment term based on the ligand size. Specifically, we add the square root of the number of ligand atoms, \(\sqrt{n^l_i}\), to the predicted radius, resulting in a final pocket radius of \(\hat{r} + \sqrt{n^l_i}\).
\begin{equation}
\mathcal{L}_r = l_{Huber}(r, \hat{r}),\quad \hat{r} = \phi_r(\sum_i h_i),
\end{equation}
where $\phi_r$ is the MLP and $h_i$ is the updated ligand atomic state in the pocket prediction module.

The total loss of pocket prediction comprises the pocket classification loss $\mathcal{L}_{cls}$, pocket centre loss $\mathcal{L}_{cen}$, and pocket radius prediction loss $\mathcal{L}_r$.
\begin{equation}
\mathcal{L}_{pocket} =\mathcal{L}_{cls} + \mathcal{L}_{cen} + \alpha_{1} \mathcal{L}_r,
\end{equation}
where $\alpha_{1}$ is the weight factor.

\subsection{Molecular Docking}\label{subsec3.5}

During the docking phase, our primary objective is to predict the coordinates of the ligand atoms when docked with the protein, based on the pocket structure \(\mathcal{G}^{p*}\) obtained earlier. Using the \(M_2\) CWFBind layers for iterative coordinate optimisation, the ligand conformation is progressively refined. As a result, the final predicted coordinates of the ligand atoms are obtained as \(\{\hat{\mathbf{x}}_i^{l} \}_{1 \leq i \leq n^l}\).

In the training phase, the docking loss $\mathcal{L}_{docking}$ consists of the coordinate loss $\mathcal{L}_{coord}$ and the distance map loss $\mathcal{L}_{dist}$.
\begin{equation}
\mathcal{L}_{docking} = \mathcal{L}_{coord} +\mathcal{L}_{dist}.
\end{equation}

The coordinate loss $\mathcal{L}_{coord}$ is computed as the Huber loss between the predicted ligand atom coordinates $\{\hat{\mathbf{x}}_i^l\}_{1 \leq i \leq n^l}$ and the corresponding ground truth coordinates $\{\mathbf{x}_i^l\}_{1 \leq i \leq n^l}$.
Accurate prediction of these coordinates is crucial, as they directly determine the ligand's docking position within the protein pocket. Minimising $\mathcal{L}_{coord}$ guides the model to better approximate the true spatial conformation of the ligand. In addition, the distance map loss $\mathcal{L}_{dist}$ is formulated as the sum of three $L_2$ losses, designed to further refine the relative spatial arrangement between protein and ligand atoms.

\begin{equation}
\begin{aligned}
\mathcal{L}_{dist} = \frac{1}{n^l n^{p*}} \Bigg[ &\sum_{i=1}^{n^l} \sum_{j=1}^{n^{p*}} (D_{ij} - \widetilde{D}_{ij})^2 + \sum_{i=1}^{n^l} \sum_{j=1}^{n^{p*}} (D_{ij} - \widehat{D}_{ij})^2+ \gamma \sum_{i=1}^{n^l} \sum_{j=1}^{n^{p*}} (\widetilde{D}_{ij} - \widehat{D}_{ij})^2 \Bigg],
\end{aligned}
\end{equation}
where $D_{ij}$  is the true distance matrix. The distance computed based on the predicted coordinates is given by $\widetilde{D}_{ij} = \|\hat{\mathbf{x}}_i^{l} - \hat{\mathbf{x}}_j^{l}\|$. The distance predicted by the pairwise embedding $\mathbf{z}_{ij}^l$ is given by $\widehat{D}_{ij} = \text{MLP}(\mathbf{z}_{ij}^l)$.

\subsection{Training Process}\label{subsec3.6}

To reduce the discrepancy between training and inference caused by the use of predicted binding pockets, a curriculum-inspired sampling strategy~\cite{pei2023fabind} is adopted. This progressive approach gradually introduces predicted pockets into the training process rather than relying solely on ground-truth annotations. By allowing the model to incrementally adapt to the uncertainty and variability associated with prediction-based inputs, it enhances robustness and improves alignment between training and inference phases.
The total training loss consists of pocket prediction loss and docking loss.
\begin{equation}
\mathcal{L} = \mathcal{L}_{pocket} + \mathcal{L}_{docking}.
\end{equation}

The pseudo-code for the CWFBind is given in Algorithm~\ref{algo1}.
\renewcommand{\algorithmicrequire}{\textbf{Input:}}
\renewcommand{\algorithmicensure}{\textbf{Output:}}
\begin{algorithm}[!t]
    \caption{CWFBind}
    \label{algo1}
 
    \begin{algorithmic}[1]
        \Require Training dataset $\mathcal{D}$, test dataset  $\mathcal{D}_t$, total epoch $T$, predicted pocket center threshold $T_p$, hyperparameter $\alpha_1$, true coordinate center $x^p_{\text{true}}$, Encoder $\mathrm{Enc}$ (LCF + ESM-2 + TorchDrug), Pocket predictor $\Phi_p$ (1 FABind layer + residue classifier), Docking predictor $\Phi_d$ (4 FABind layers + 8 refinement iterations), and Optimizer $\mathrm{Opt}$ (AdamW, $\eta = 5\text{e-5}$).
        \Ensure Predicted ligand coordinates $\hat{x}^l_t$
        
        \State $\mathcal{G}, \mathcal{G}^l, \mathcal{G}^p \gets \mathrm{Enc}(\mathcal{D})$
        \For{each epoch $\in \{1,2,\ldots, T\}$}
            \State $\mathcal{G}^l, \mathcal{G}^p, \hat{y} \gets \Phi_p(\mathcal{G})$
            \State Calculate the center $\hat{x}^p$ of the pocket using Eq. (8)
            \State Calculate losses $\mathcal{L}_{\text{cls}}$, $\mathcal{L}_{\text{cen}}$, and $\mathcal{L}_{\text{r}}$ using Eqs. (7), (9), and (10)
            \State The predicted pocket radius $\hat{R} = \hat{r} + \sqrt{n^l_i}$ is obtained according to Eq. (10)
            \State $\mathcal{L}_{\text{pocket}} \gets \mathcal{L}_{\text{cls}} + \mathcal{L}_{\text{cen}} + \alpha_{1} \mathcal{L}_r$
            \If{$\text{epoch} < T_p$}
                \State $x^p_{\text{use}} \leftarrow x^p_{\text{true}}$
            \Else 
                \State $x^p_{\text{use}} \leftarrow \hat{x}^p$
            \EndIf
            \State $\hat{x}^l \leftarrow \Phi_d(\mathcal{G}^l, \mathcal{G}^p, x^p_{\text{use}}, \hat{R})$
            \State Calculate loss $\mathcal{L}_{\text{dist}}$ using Eq. (13)
            \State $\mathcal{L}_{\text{coord}} \leftarrow l_{\text{Huber}}(x_i^l,\hat{x}_i^l)$
            \State $\mathcal{L}_{\text{docking}} \leftarrow \mathcal{L}_{\text{dist}} + \mathcal{L}_{\text{coord}}$
            \State $\mathcal{L} \leftarrow \mathcal{L}_{\text{pocket}} + \mathcal{L}_{\text{docking}}$
            \State Optimize loss $\mathcal{L}$ through $\mathrm{Opt}$
        \EndFor
        \State $\hat{x}^l_t \leftarrow \Phi_d(\Phi_p(\mathrm{Enc}(\mathcal{D}_t)))$
    \end{algorithmic}
\end{algorithm}

\section{Experiments and Results}\label{sec4}

Experiments are performed on the PDBbind v2020 dataset~\cite{burley2021rcsb}, a widely used dataset for molecular docking problems containing 19,443 protein-ligand complex structures. For fair comparison with prior work, the experiments follow the same data splitting and preprocessing protocols as those used in TankBind~\cite{lu2022tankbind} and FABind~\cite{pei2023fabind}. First, complexes from PDBbind v2020 that cannot be processed by RDKit or TorchDrug are excluded. Second, to address the issue of multiple equivalent binding sites due to receptor symmetry, protein chains with no atoms within $10 \text{\AA}$ of any ligand atom are removed. Finally, complexes are filtered out if they contain five or fewer contacts (within $10 \text{\AA}$) between protein $C_\alpha$ atoms and ligand atoms, or if the number of ligand atoms is greater than or equal to 100.


To evaluate the performance of the proposed CWFBind method, we compare it against representative docking approaches across three categories: (1) traditional molecular docking software, including GLIDE~\cite{friesner2004glide}, VINA~\cite{trott2010autodock}, SMINA~\cite{koes2013lessons}, and GNINA~\cite{mcnutt2021gnina}; (2) deep learning-based regression methods, such as EquiBind~\cite{stark2022equibind}, TankBind~\cite{lu2022tankbind}, E3Bind~\cite{zhang2023e3bind}, and FABind~\cite{pei2023fabind}; and (3) deep learning-based sampling methods, including DiffDock~\cite{corso2023diffdock} and FABind+~\cite{gao2025fabind+}.

We select two evaluation metrics for the predicted ligand binding pose. One is the Ligand Root Mean Square Deviation (LRMSD), which measures the deviation between predicted and true ligand coordinates. Another is Centroid Distance (CD), which calculates the Euclidean distance between the predicted ligand structure and the centre of mass of the true ligand structure. The calculation formula is as follows:
\[
\text{RMSD} = \sqrt{\frac{1}{n^l} \sum_{i=1}^{n^l} (\hat{\mathbf{x}}^l_i - \mathbf{x}^l_i)^2},
\]
\[
\text{CD} = \sqrt{  (\hat{\mathbf{x}}^{p} - \mathbf{x}^{p})^2},
\]
where $n^l$ denotes the total number of atoms in the ligand molecule. The $\hat{\mathbf{x}}^l_i$ and $\mathbf{x}^l_i$ denote the predicted and true coordinates of the $i$-th ligand atom, respectively. \(\hat{\mathbf{x}}^{p}\) and \({\mathbf{x}}^{p}\) represent the predicted and real ligand centroid coordinates, respectively.


One CWFBind layer and four CWFBind layers are selected for the pocket prediction and molecular docking modules, respectively, with the number of hidden layers set to 128 and 512. The CWFBind model is trained with around 450 epochs using the AdamW optimiser on an NVIDIA GeForce RTX 3080 Ti 12 GB GPU with a batch size of 3. The PyTorch framework was used for the experiments. During training, the model's scheduler is LinearLR, and the learning rate is 5e-5. The weight factor $\alpha_{1}=0.05$ in the loss function.

 
\subsection{Comparison with baselines}

In the blind flexible self-docking task, where the protein structure is known but the binding site (ligand conformation) is unknown, the objective is to predict the three-dimensional conformation of the ligand-binding pose. Table~\ref{table1} shows the results for all test sets, with the best results in bold and sub-optimal results highlighted by horizontal lines. CWFBind performs best or second best in most of the metrics. Specifically, CWFBind significantly outperforms traditional docking tools such as GLIDE, VINA, SMINA, and GNINA in both prediction accuracy and computational efficiency. This advantage stems from the limitations of conventional methods, which depend on predefined, rigid binding pockets and simplified physical force fields, resulting in biased pocket localization and inaccurate energy estimations that fail to capture the dynamic and complex nature of biological interactions. For the deep learning based regression method, CWFBind outperforms EquiBind, TankBind, E3Bind, and FABind under both LRMSD and CD metrics. The means for LRMSD and CD are improved by $8.5 \%$ and $27.0 \%$, respectively, compared to the means for FABind. These results demonstrate that the introduced LCF graph structure strengthens the ability to characterise irregular binding cavities. Additionally, the dynamic radius associated with the ligand structure accurately identifies the pocket position of the protein. In terms of sampling methods, CWFBind performs better than DiffDock, a molecular docking method based on the diffusion model, and performs slightly below FABind+ for the percentage of LRMSD and CD metrics less than $5 \text{\AA}$. This is because the conformations generated by the diffusion model may violate physical constraints, such as bond lengths and angles. FABind+ clusters pockets using the DBSCAN method~\cite{braun2022mapping} and explores different binding sites using a dropout-based method to generate conformations. A common problem with sampling methods is their high computational cost. Overall, our method remains highly competitive, greatly enhancing the efficiency of training and prediction while maintaining the accuracy of molecular docking.

\begin{table*}[t]
\caption{\\Comparative results of blind and flexible self-docking.}
\label{table1}
\tabcolsep=0pt
\begin{tabular*}{\textwidth}{@{\extracolsep{\fill}}lccccccccccccc@{\extracolsep{\fill}}}
\toprule
& \multicolumn{6}{c}{\textbf{Ligand RMSD}} & \multicolumn{6}{c}{\textbf{Centroid Distance}} & \textbf{Average} \\
\cmidrule(lr){2-7} \cmidrule(lr){8-13} 
&\multicolumn{4}{c}{Percentiles $\downarrow$}  & \multicolumn{2}{c}{$\%$ Below $\uparrow$} & \multicolumn{4}{c}{Percentiles $\downarrow$} & \multicolumn{2}{c}{$\%$ Below $\uparrow$} & \textbf{Runtime} \\
\cmidrule(lr){2-5} \cmidrule(lr){6-7} \cmidrule(lr){8-11} \cmidrule(lr){12-13}
\textbf{Method} & 25\% & 50\% & 75\% & Mean & 2\AA & 5\AA & 25\% & 50\% & 75\% & Mean & 2\AA & 5\AA & \textbf{(s)} \\
\midrule
GNINA & 2.8 & 8.7 & 22.1 & 13.3 & 21.2 & 37.1 & 1.0 & 4.5 & 21.2 & 11.5 & 36.0 & 52.0 & 146 \\
SMINA & 3.8 & 8.1 & 17.9 & 12.1 & 13.5 & 33.9 & 1.3 & 3.7 & 16.2 & 9.8 & 38.0 & 55.9 & 146* \\
GLIDE & 2.6 & 9.3 & 28.1 & 16.2 & 21.8 & 33.6 & 0.8 & 5.9 & 26.9 & 14.4 & 36.1 & 48.7 & 1405* \\
VINA & 5.7 & 10.7 & 21.4 & 14.7 & 5.5 & 21.2 & 1.6 & 6.2 & 20.1 & 12.1 & 26.5 & 47.1 & 205* \\
\midrule
EquiBind & 3.8 & 6.2 & 10.3 & 8.2 & 5.5 & 39.1 & 1.3 & 2.6 & 7.4 & 5.6 & 40.0 & 67.5 & \textbf{0.03} \\
TankBind & 2.6 & 4.2 & 7.6 & 7.8 & 17.6 & 57.8 & 0.8 & 1.7 & 4.3 & 5.9 & 55.0 & 77.8 & 0.87 \\
E3Bind & 2.1 & 3.8 & 7.8 & 7.2 & 23.4 & 60.0 & 0.7 & 1.5 & 4.0 & 5.1 & 60.0 & 78.8 & 0.44 \\
DiffDock (10) & 1.5 & 3.6 & 7.1 & - & 35.0 & 61.7 & \underline{0.5} & 1.2 & 3.3 & - & 63.1 & 80.7 & 20.81 \\
DiffDock (40) & 1.4 & 3.3 & 7.3 & - & 38.2 & 63.2 & \underline{0.5} & 1.2 & 3.2 & - & 64.5 & 80.5 & 82.83 \\
FABind & 1.7 & 3.1 & 6.7 & 6.4 & 33.1 & 64.2 & 0.7 & 1.3 & 3.6 & 4.7 & 60.3 & 80.2 &  0.12 \\
FABind+ & \textbf{1.2} & \underline{2.6} & 5.8 & \textbf{5.2} & \underline{43.5} & 71.1 & \textbf{0.4} & \textbf{1.0} & 2.9 & \textbf{3.5} & \textbf{67.5} & \textbf{84.0} & 0.16 \\
FABind+(10) & \underline{1.3} & 2.7 & \textbf{5.4} & \textbf{5.2} & 42.4 & \underline{71.6} & \underline{0.5} & \underline{1.1} & 2.8 & \textbf{3.5} & \underline{67.8} & 84.6 & 1.6 \\
FABind+(40) & \textbf{1.2} & \textbf{2.4} & \underline{5.6} & \textbf{5.2} & \textbf{44.9} & 71.3 & \underline{0.5} & \textbf{1.0} & \textbf{2.7} & \textbf{3.5} & \textbf{68.3} & \underline{85.2} & 6.4 \\
\midrule
CWFBind & 1.4 & 2.8 & \textbf{6.1} & 5.9 & 38.3 & \textbf{71.9} & 0.6 & \underline{1.1} & \textbf{2.7} & \underline{3.7} & 66.0 & \textbf{85.8} & \underline{0.09} \\
\bottomrule
\end{tabular*}

\noindent Note: The symbol * indicates CPU results. Bold values represent the best performance, and underlined values represent the second-best performance. The number of molecular conformations sampled by DiffDock is indicated in parentheses.
\end{table*}


To further validate the generalisation ability of our method to completely novel protein structures, we added a filtering step to retain only the 144 protein samples not used in the training and validation sets. The experimental results are shown in Table~\ref{table2}. As can be seen, CWFBind performs well, outperforming all existing molecular docking methods except FABIND+. Among regression deep learning docking methods, our approach generalises well and is robust to unknown proteins, demonstrating the coverage capabilities of the dynamic radius boosting pocket. The introduction of degree weights in the messaging process enables more accurate prediction of binding sites, which is a key step in advancing our understanding of these processes. This is particularly evident in the CWFBind rankings of $50 \%$ and less than $5 \text{\AA}$ accuracy, with CD scores of 1.4 and $77.2 \%$, respectively, higher than those of FABind+ (40). CWFBind achieves an accuracy of $63.9 \%$ at LRMSD of less than $5 \text{\AA}$, implying that the ligand can be found in the correct conformation at the atomic level. Although CWFBind's performance is lower than that of FABind+ in the less than $2\text{\AA}$ metric, this is due to the optimisation goal of the FABind+ confidence model, which primarily focuses on optimising the ligand conformation to achieve less than $2 \text{\AA}$ accuracy.

\begin{table*}[t]
\caption{\\Comparative results of blind flexible self-docking on unseen receptors.}
\label{table2}
\tabcolsep=0pt
\begin{tabular*}{\textwidth}{@{\extracolsep{\fill}}lccccccccccccc@{\extracolsep{\fill}}}
\toprule
& \multicolumn{6}{c}{\textbf{Ligand RMSD}} & \multicolumn{6}{c}{\textbf{Centroid Distance}}  \\
\cmidrule(lr){2-7} \cmidrule(lr){8-13} 
&\multicolumn{4}{c}{Percentiles $\downarrow$}  & \multicolumn{2}{c}{$\%$ Below $\uparrow$} & \multicolumn{4}{c}{Percentiles $\downarrow$} & \multicolumn{2}{c}{$\%$ Below $\uparrow$} \\
\cmidrule(lr){2-5} \cmidrule(lr){6-7} \cmidrule(lr){8-11} \cmidrule(lr){12-13}
\textbf{Method} & 25\% & 50\% & 75\% & Mean & 2\AA & 5\AA & 25\% & 50\% & 75\% & Mean & 2\AA & 5\AA \\
\midrule
GNINA & 4.5 & 13.4 & 27.8 & 16.7 & 13.9 & 27.8 & 2.0 & 10.1 & 27.0 & 15.1 & 25.7 & 39.5 \\
SMINA & 4.8 & 10.9 & 26.0 & 15.7 & 9.0 & 25.7 & 1.6 & 6.5 & 25.7 & 13.6 & 29.9 & 41.7 \\
GLIDE & 3.4 & 18.0 & 31.4 & 19.6 & 19.6 & 28.7 & 1.6 & 17.6 & 29.1 & 18.1 & 29.4 & 40.6 \\
VINA & 7.9 & 16.6 & 27.1 & 18.7 & 1.4 & 12.0 & 2.4 & 15.7 & 26.2 & 16.1 & 20.4 & 37.3 \\
\midrule
EquiBind & 5.9 & 9.1 & 14.3 & 11.3 & 0.7 & 18.8 & 1.2 & 6.3 & 12.9 & 8.9 & 16.7 & 43.8 \\
TankBind & 3.4 & 5.7 & 10.8 & 10.5 & 3.5 & 43.7 & 1.2 & 2.6 & 8.4 & 8.2 & 40.9 & 70.8 \\
E3Bind & 3.0 & 6.1 & 10.2 & 10.1 & 6.3 & 38.9 & 1.2 & 2.3 & 7.0 & 7.6 & 43.8 & 66.0 \\
DiffDock (10) & 3.2 & 6.4 & 16.5 & 11.8 & 14.2 & 38.7 & 1.1 & 2.8 & 13.3 & 9.3 & 39.7 & 62.6  \\
DiffDock (40) & 2.8 & 6.4 & 16.3 & 12.0 & 17.2 & 42.3 & 1.0 & 2.7 & 14.2 & 9.8 & 43.3 & 62.6  \\
FABind & 2.2 & 3.4 & \textbf{8.3} & 7.7 & 19.4 & 60.4 & 0.9 & \underline{1.5} & 4.7 & 5.9 & 57.6 & 75.7  \\
FABind+ & \textbf{1.6} & 3.3 & 8.9 & \textbf{7.0} & \underline{34.7} & \underline{63.2} & \textbf{0.5} & \underline{1.5} & \textbf{4.2} & \textbf{5.1} & \underline{58.3} & \underline{77.1} \\
FABind+(10) & \textbf{1.6} & \underline{3.2} & 9.0 & 7.4 & 33.3 & 61.8 & \underline{0.6} & \textbf{1.4} & \underline{4.3} & 5.7 & \textbf{59.0} & 75.0 \\
FABind+(40) & \textbf{1.6} & 3.3 & 8.8 & \underline{7.1} & \textbf{35.4} & 61.1 & \underline{0.6} & \underline{1.5} & 4.9 & \underline{5.3} & \underline{58.3} & 76.3  \\
\midrule
CWFBind & \underline{1.7} & \textbf{3.1} & \underline{8.6} & 7.5 & 32.6 & \textbf{63.9} & 0.8 & \textbf{1.4} & \underline{4.3} & 5.7 & 57.6 & \textbf{77.2} \\
\bottomrule
\end{tabular*}
\noindent Note: Bold values represent the best performance, and underlined values represent the second-best performance. The number of molecular conformations sampled by DiffDock is indicated in parentheses.
\end{table*}

\subsection {Ablation study}

We conducted an ablation study to systematically assess the contribution of each core component in CWFBind. As shown in Table~\ref{table3}, removing the local curvature-based features ("w/o LCF"), degree-based weighting ("w/o weight"), dynamic pocket radius adjustment ("w/o dynamic radius"), and the class-balanced focal loss ("w/o loss") each resulted in a notable decline in performance. In contrast, the complete model (last row) achieved the best results across all evaluation metrics. Notably, removing the LCF features had the most substantial impact, with the proportion of predictions achieving ligand RMSD$<2\text{\AA}$ dropping from $38.3\%$ to $30.0\%$. This highlights the critical role of incorporating local geometric complementarity between proteins and ligands for accurate binding conformation prediction. Without such geometric information, the model's capacity to capture critical spatial alignment cues is significantly impaired. The implementation of degree node weights has been demonstrated to engender an enhancement in the overall ligand RMSD metrics, thereby indicating that the aggregation of neighbourhood information based on the degree of connectivity of the nodes in the protein-binding pocket can more efficaciously capture the topological influence of key residues. Following the implementation of dynamic radius adjustment, there was an enhancement in the ligand RMSD$<5\text{\AA}$ accuracy from $68.6\%$ to $71.9\%$. This is primarily due to the substantial improvement in binding pocket recognition accuracy by adaptive pocket radius. After replacing the original binary classification cross-entropy loss, the molecular docking accuracy is improved by $15.7\%$ and $4.4\%$, respectively, which effectively solved the data imbalance problem.

\begin{table}[!t]
\caption{\\Ablation study.}%
\label{table3}
\begin{tabular*}{\columnwidth}{@{\extracolsep\fill}lccc@{\extracolsep\fill}}
\toprule

\multirow{2}{*}{\textbf{Methods}} & \textbf{RMSD} & \textbf{RMSD} & \textbf{RMSD} \\
                                     & \textbf{Mean \AA  $\downarrow$} & \textbf{$<$ 2\AA (\%)  $\uparrow$} & \textbf{$<$ 5\AA(\%) $\uparrow$} \\
\midrule
w/o. LCF & 6.0 & 30.0 & 68.6 \\
w/o. weight & 5.9 & 36.4 & 71.1 \\
w/o. dynamic radius & 6.1 & 36.6 & 68.6 \\
w/o. loss & 5.9 & 33.1 & 68.9 \\
CWFBind & \textbf{5.7} & \textbf{38.3} & \textbf{71.9} \\
\bottomrule
\end{tabular*}
\end{table}

\subsection {Inference Efficiency}

Molecular docking efficiency is a core metric for the assessment of the utility of the method. The mean time (s) required for each sample is enumerated in Table~\ref{table1}, with * denoting CPU results. The inference time of the CWFBind method is 0.09s, which is the second-fastest of the methods under comparison. Despite the marginal increase observed in comparison with EquiBind 0.07s, it has been determined that EquiBind exhibits a pronounced decrease in ligand docking accuracy when evaluated in conjunction with the accuracy data presented in Tables~\ref{table1} and ~\ref{table2}. The conventional docking tools GLIDE, Vina, SMINA, and GNINA have been identified as the least efficient (requiring $\geq 146s$), thus emphasising the revolutionary acceleration of the molecular docking process by deep learning. Among the end-to-end deep learning methods, CWFBind demonstrates significant efficiency advantages. The present study demonstrates that the CWFBind algorithm is 71 times faster than FABind+ when 40 conformational samples are used as the basis for comparison. This is due to the single forward prediction architecture of CWFBind, which completely circumvents the sampling and scoring process. Furthermore, CWFBind is 690 times faster than DiffDock, since it eliminates the diffusion generation process and the external pocket prediction module. CWFBind has achieved a synergistic breakthrough in accuracy and efficiency through a streamlined end-to-end architecture that compresses the docking consumption time to the sub-second level while maintaining high-precision prediction capability. This provides a technological foundation for large-scale virtual screening and real-time drug discovery and development.

\subsection {Case Studies}
\begin{figure}[t]%
    \centering 
    \includegraphics[width=\textwidth]{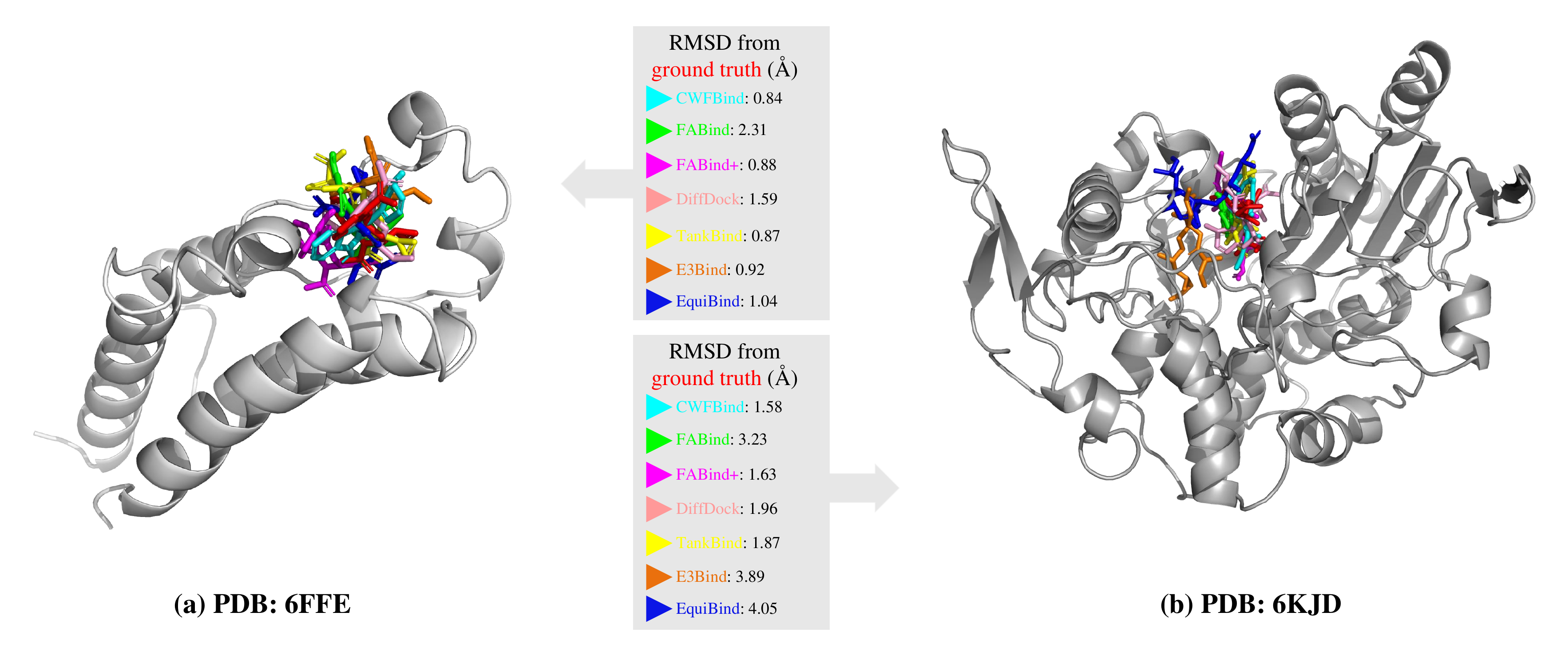}
    \caption{ Ligand position prediction of proteins at unknown binding sites by different methods, with true position (red), CWFBind (cyan), FABind (green), FABind+ (megenta), DiffDock (pink), TankBind (yellow), E3Bind (orange), and EquiBind (blue). (a) PDB 6FFE (b) Unseen protein PDB 6KJD.} 
    \label{figure3} 
\end{figure}

To further demonstrate the docking performance of CWFBind, visualizations are generated for selected cases from the test set. Fig.~\ref{figure3}(a) presents the predicted ligand positions on protein 6FFE using different methods, while Fig.~\ref{figure3}(b) shows predictions for protein 6KJD, which is excluded from the training and validation sets. As shown in Fig.~\ref{figure3}(a), all seven methods converge on the binding pocket region of protein 6FFE. Among them, CWFBind achieves the highest accuracy, aligning with the ground-truth ligand pose at an RMSD of just $0.84\text{\AA}$. TankBind and FABind+ are sub-optimal with RMSDs of 0.87 and 0.88, respectively. FABind performs the worst of the three, possessing the highest RMSD. This may be attributed to the fact that FABind lacks saliency modelling of geometric features, and the pocket prediction module relies on the extraction of features from a spherical region of a fixed radius, which is unable to dynamically adapt to pocket depths and shapes of different target sites. As illustrated in PDB 6KJD (Fig.~\ref{figure3}(b)), the binding pockets identified by EquiBind and E3Bind in PDB 6KJD deviate from the correct position, while all other methods accurately identify these binding pockets. Although FABind correctly identifies the position, a deviation in the conformational pose is observed, resulting in a substantial RMSD. In contrast, CWFBind demonstrates an ability to accurately predict binding poses, as evidenced by its ability to obtain a lowest RMSD of $1.58 \text{\AA}$. This finding suggests that the method exhibits satisfactory generalisation performance.

\section {Conclusion}\label{sec5}

In this study, an efficient docking method based on LCF, CWFBind, is proposed. The method achieves accurate prediction of ligand binding conformations through three stages: feature extraction, binding pocket prediction, and coordinate refinement. Firstly, integrates geometric, chemical, sequence, and structural features to comprehensively represent protein and ligand properties. Pocket prediction is then performed using CWFBind layers, while final ligand poses are refined through an iterative coordinate optimisation strategy. The experimental results demonstrate that CWFBind boasts significant advantages in terms of prediction accuracy and computational efficiency, particularly in the context of blind docking scenarios. Compared to existing sampling-based and regression-based methods, CWFBind shows clear advantages in generalisation and reliability. Nevertheless, the current framework is restricted to semi-flexible protein–ligand docking and does not yet account for conformational changes in protein structures. In future work, we plan to extend the model to fully flexible docking scenarios, enabling more accurate simulations of dynamic binding processes.





\bibliographystyle{elsarticle-num}
\bibliography{refs.bib}

\end{document}